# Single Document Extractive Summarization using Domination in Hypergraph

Aamir Miyajiwala[a], Aabha Pingle[a], Sheetal Sonawane[a] and Surajit Kr. Nath[b]

*[a]Department of Computer Engineering, SCTR's Pune Institute of Computer Technology; [b]Bodoland University, Assam, India*

Abstract

Background:
Automatic Text Summarization (ATS) in Natural Language Processing has been an important task in Information Retrieval. It compresses a document to create a summary that captures all the relevant and important information conveyed in the document. This study explores Hypergraph for extractive text summarization of single documents.
Objective: This study explores a novel method of leveraging the property of domination in hypergraphs to generate an extractive summary and compare its performance with state of the art graphbased methods.
Method: Our work aims to generate an extractive summary by creating a sentence hypergraph where each sentence represents a node and the edge is a keyword or a named entity that contains the sentences in which it occurs. We generate a hypergraph where each edge is a keyword or an important topic and the nodes are sentences containing those keywords. Then we apply a greedy algorithm to find the dominating set of the hypergraph which will contain sentences that will form the extractive summary.

Results: Based on the ROUGE metrics calculated for different datasets, our approach gives encouraging results which are comparable to traditional methods of TextRank and LexRank. We have evaluated the performance of our approach against human summaries on benchmark datasets using the ROUGE toolkit.

Conclusion: Using the property of domination in hypergraph we generated extractive summaries of single documents with decent ROUGE scores. Future work involves the use of co-reference resolutionas part of the pipeline.

## 1. INTRODUCTION

Recently with the rise in big data, automatic text summarization of documents has attracted widespread interest. Furthermore, existing text summarizers aremostly extractive in nature, i.e. they produce summaries that contain the most informative sentences from the input text. Particularly graph based approaches are very popular [1, 2] and have been widely adopted for this task. This is because they are able to naturally connect the entities between sentences and capture complex relationships. The nodes are used to represent sentences and the similarities between the sentences are shown by edges in conventional graph-based approaches. Once the graph is created, they are ranked using popular algorithmssuch as LexRank[3] and TextRank [4] and the summary includes the first n ranked sentences. But since every edge only connects a pair of nodes, this technique misses the group relations between sentences. To alleviate this critical issue, hypergraphs were proposed [5, 6].

Here each hyperedge contains a cluster of lexically similar sentences known as themes. However, this method of constructing a hypergraph fails to account for the possibility of overlapping of themes, where a sentence carries multiple information and therefore could be a part of more than one theme. In this paper we present a sentence hypergraph for single document summarization where we extract key-words and entities from the document and use them as a hyperedge to group sentences that contain those keywords and entities. This increases topical diversity as a sentence can contain more than one entity or keyword and hence be part of more than one hyperedge. Once the hypergraph is created, we extract sentences that contain important keywords and are representative of the entire document. This is analogous to finding the dominating set of a hypergraph. Our work is one of the first to leverage domination in hypergraphs for unsupervised extractive text summarization.

The paper is organized as follows. We discuss the relevant literature in section 2. Section 3 talks about Hypergraphs and introduces the concept of Domination. In section 4, we present our methodology in detail. The experimental findings are then displayed in section 5 along with example sentence hypergraphs. Section 6 concludes with a discussion and final remarks.

## 2. LITERATURE REVIEW

Extractive summarization models are categorized into two groups based on the mechanism used to determine important sentences: feature-based and graph-based methods. A standard procedure for finding extractive summaries has been using various graph-based data structures. In graph-based summarizers, the graph is made using sentences in the documents as nodes and edges representing the similarity links between these sentences. Then key sentences are found out either via node ranking algorithms or global optimization strategies. For node ranking algorithms, a saliency score is computed for each sentence and then the summaries are constructed using highly scored sentences. Over the years, researchers have come up with a number of algorithms for sentence scoring [3, 7]. But these systems based on node ranking have the disadvantage of computing the saliencyscore for individual sentences only. They overlook how these particular sentences are related to one another. Hence, a number of greedy algorithms were used for findinga set of jointly relevant sentences [8, 9]. These methodspropose approximation algorithms in order to solve therespective NP-hard problems. In one such attempt [9], theauthors have constructed a sentence graph where vertices represent sentences and edges are constructed between twovertices if they are similar. Sentences are represented asvectors based on tf-isf, and the cosine similarity for each pairof sentences determines the presence of an edge betweenthem. They have made use of graph domination to describethe summary. The paper proposes approximate algorithms for finding the minimum dominating set of the sentence graph.

As an alternative to finding semantic similarities between a pair of sentences using simple graphs, hypergraph models have been used by a number of researchers for text modeling [5, 10]. Hyperedges are used to represent a group of sentences having some common topical link between them. It was observed that these clusters do not overlap effectively and that the hyperedges do not have sufficient common nodes. These methods often suffer from their high computational complexity and storage cost.

Reportedly, domination in hypergraphs has never before been used for finding the text summary of a single document.

## 3. BASIC TERMINOLOGIES

### 3.1. Hypergraph

A simple graph is extended into a hypergraph, where any number of vertices can be connected by a single edge [11] . Two sets X and E, which are referred to as the vertex set and edge set, make up the hypergraph H where each member of $E$ is a non-empty subset of $X$ and which satisfies the following condition:

$$\bigcup_{i=1}^{m} Ei = X$$

Where $E_i$ are members of $E$. For example: Let the vertex set be $V = \{v_1,v_2,v_3,v_4,v_5,v_6,v_7\}$ and the edge set be $E = \{E_1,E_2,E_3,E_4,E_5\}$ of the hypergraph like in fig.1

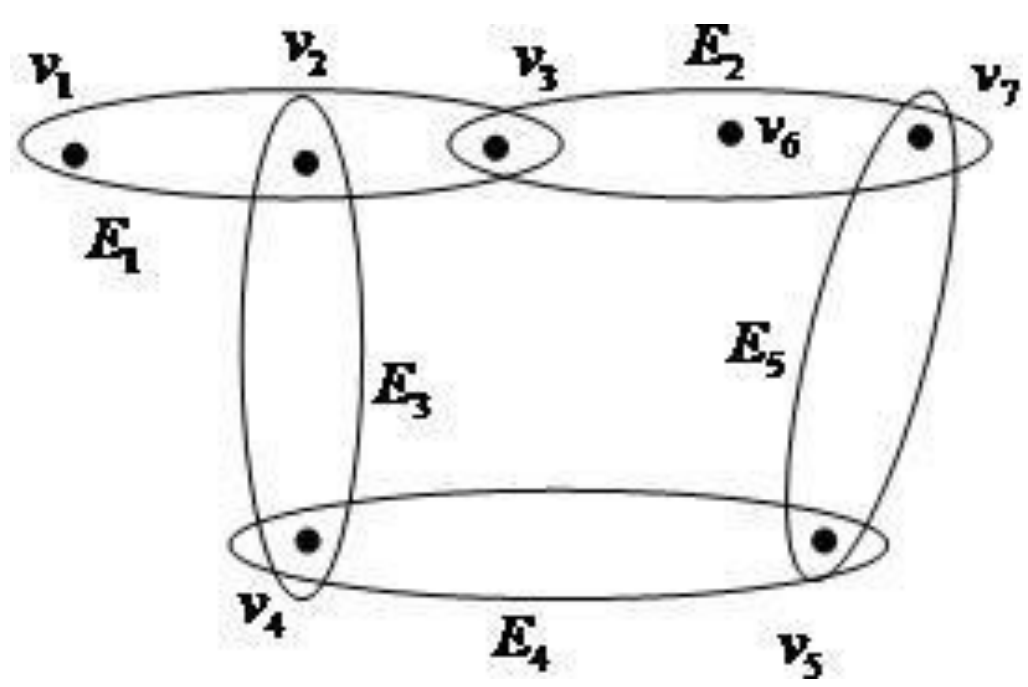

**Fig. 1: Example of Hypergraph**

Any two vertices are said to be adjacent if they have at least one common edge. Instead, we can say that all vertices included within a single edge are adjacent to one another. If any two edges share at least one vertex, they are considered to be adjacent.

### 3.2. Domination

By B.D. Acharya, the idea of domination in hypergraphs was first introduced [12]. Before explaining about domination, we need to describe some other concepts as follows:

A set of vertices in a hypergraph $H$ that does not have any edges $E$ with $|E|>1$ is known as the stable set $S$. The stability number of $H$ is the maximum cardinality of a stable set. It is denoted by $\alpha(H)$. The independent set is the collection of vertices such that no two vertices are adjacent in $H$. The independence number of $H$ which is denoted by $\beta(H)$ is the maximum cardinality of an independence set. Clearly since the set $S_\alpha(H)$ of all maximal stable sets in $H$ contains the set $S_\beta(H)$ of all maximal independence set in $H$ it follows that, for any hypergraph $H$, $\alpha(H)\geq\beta(H)$. [12]

***Definition 1***.[12] Consider a hypergraph to be $H = (X, E)$. A set $D\subseteq X$ is a dominating set of $H$'s if for *every* $x\subset X - D$ there exists $u\subset D$ such that $v$ and $u$ are adjacent in $H$, that is they are contained in some $e$, where $e\subset E$.

For example if we consider the set $D=\{v_1,v_2,v_3\}$ in the hypergraph shown in fig1. This set is a dominating set ofthat hypergraph. Dominating set with least cardinality is called **minimum dominating set** and the cardinality of this set is the **domination number**, denoted by $\gamma(H)$.

***Theorem 1.*** [12] Consider a hypergraph to be $H = (X, E)$. All maximum stable sets are thus minimal dominating sets of $H$.

## 4. METHODOLOGY

This system architecture is shown in fig2. Following steps were carried out to generate the document summary.

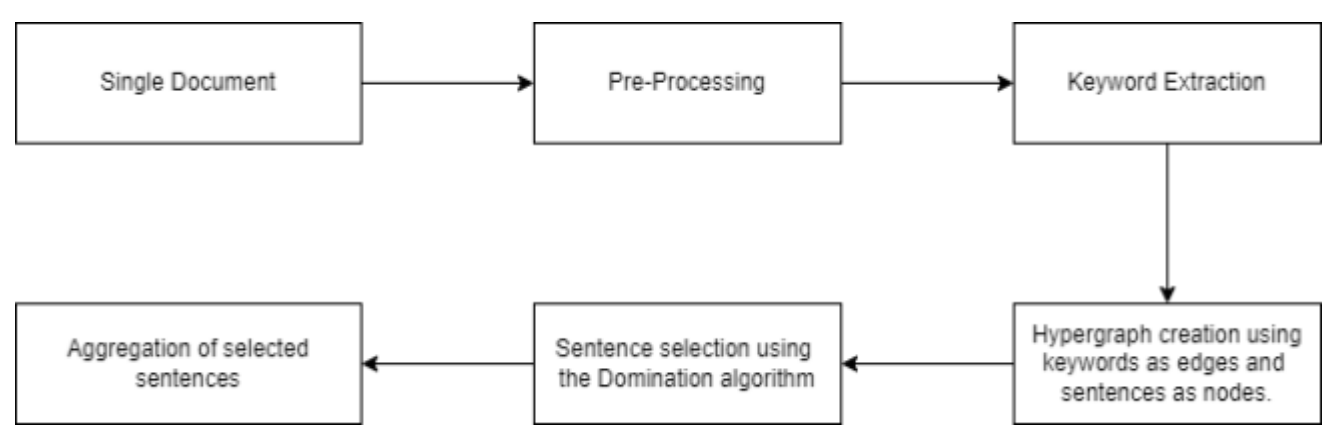

**Fig. 2: System architecture**

1. Pre-processing of the document.
2. Keywords Extraction and Entity Recognition.
3. Hypergraph creation.
4. Sentence selection using Domination.

### 4.1. Pre-processing of the document

The pre-processing steps are as follows:

1. Stopword removal using the nltk library.
2. Using wordnet lemma for lemmatization to normalize the text.
3. Removal of special characters and punctuations(except for the period).
4. Sentence segmentation using regular expression. Regular expression is used for this purpose to split the sentence only if the period/fullstop denotes the end of sentence.

### 4.2. Keywords Extraction and Entity Recognition

Keywords and Entity extraction is done in the following ways:

#### *4.2.1. YAKE*

Yet another Keyword Extractor (YAKE) is an unsupervised, automatic tool for keyword extraction [13]. To choose significant keywords, it analyses text statistical features from individual documents. Since it is unsupervised, no training corpus is necessary unlike tf-idf and it is language anddomain independent. Since our work deals with single document summarization where each document is independent of the other, YAKE proves to be the most suitable keyword extraction algorithm. We use the YAKE open sourcepython package for our work. The n-grams parameter was setto 1 for extracting single-word keywords.

#### *4.2.2. NER*

Named Entity Recognition has been implemented to extract entities such as people, organizations, location etc. that would ensure that the summary generated contained most of the important and relevant information. NER can be used to extract those keywords that may not have been extracted by YAKE and therefore complement each other. This is one of the primary reasons behind its use for our purpose. For implementing NER the spaCy [14] package was used.

### 4.3. Hypergraph creation.

From the keywords and entities extracted with models described in subsection 4.2, we create a list of distinct keywords. These would act as hyperedges containing sentences in which the keyword is present. The hypergraph is structured as a node as a keyword and edge is for list of sentences having that keyword. Algorithm defined in section 4.5.1 has been used for creating the hypergraph. HyperNetX Python package [15] with dictionary has been used for further processing.

### 4.4. Sentence selection using Hypergraph.

Finding the dominating set of a graph is an NP-Complete problem and therefore, a greedy algorithm described in section 4.5.2 has been implemented to find the approximate dominating set of the hypergraph. An advantage of using domination as described in Def.1 is that if a hyperedge contains only one node or sentence, that sentence will still be part of the dominating set and in turn be part of the summary. This makes sure that specific and important sentences are also included, enabling maximum topical coverage.

### 4.5. Algorithms and Program Codes.

#### *4.5.1. Algorithm for Hypergraph creation.*

**INPUT:** Keyword array (K), List of cleaned separated sentences from the documents.
**OUTPUT**: Keyword Sentence Dictionary (W).

1. **for** word in K **do**
2. li = [] *>sentences containing word*
3. **for** sentence in S **do**
4. result = findall(word, sentence)
5. **if** length (result > 0) **then** *>if word is in sentence*
6. li. append (sentence) *>append sentence to list*
7. **end if**
8. W[word] = li
9. **end for**
10. **end for**
11. return W

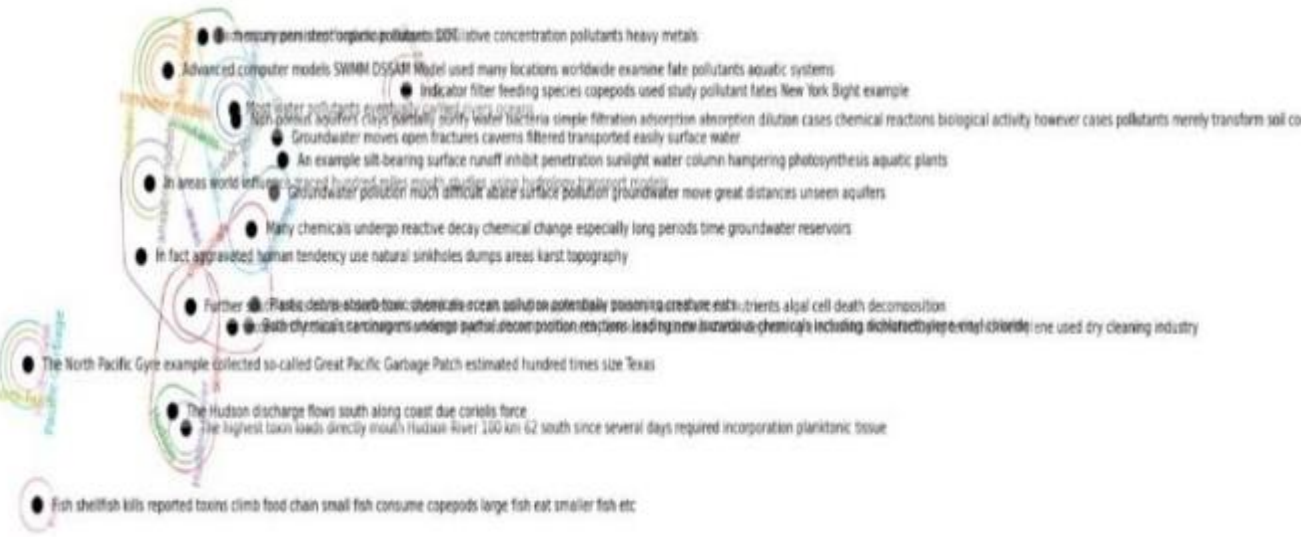

**Fig. 3: Keyword Sentence Hypergraph**

### *4.5.2. Algorithm to find dominating set.*

INPUT: Word Sentence Dictionary (W)
OUTPUT: Dominating Set (S)

1. h = Hypergraph(W) >*hypergraph object*
2. **for** i in h.nodes **do**
3. li.append(h.neighbours(i)) >*neighbours of each node*
4. **end for**
5. c=0
6. **for** i in list(h.nodes) **do**
7. mapp[i] = c >*mapping nodes to numbers*
8. c = c+1
9. **end for**
10. **for** i in range length(li) **do**
11. deg[i] = length(li[i]) >*find degree of every node*
12. **end for**
13. deg =dict(sorted(deg.items(), key=lambda item: item[1], reverse = True))
14. ver = length(li)
15. visit = np.zeros(ver) >*array initialized with zeros*
16. **for** i in deg.keys() **do**
17. **if** visit[i] == 0 **then**
18. S.append(i)
19. visit[i] == 1
20. **for** j in range(length(li[i])) do
21. if visit[mapp[li[i][j]]] == 0 then
22. visit[mapp[li[i][j]]] = 1
23. **end if**
24. **end for**
25. **end if**
26. **end for**

## 5. EXPERIMENTATION AND RESULTS

In this section, we discuss the results of our approach on different single document text summarization benchmark datasets such as DUC2002, ESSG dataset [16] which consists of 100 self-generated (Sonawane 2018) documents which were collected from Wikipedia, and the CNN/Daily Mail news [17] dataset. These datasets contain 500,100 and 25 documents respectively. The reference summaries for DUC2002 and CNN/Daily Mail news datasets are abstractive whereas the summaries for ESSG dataset are extractive. On applying the proposed algorithm, a sentence hypergraph was obtained as shown in the fig 3. ROUGE-N(Recall-Oriented Understudy for Gisting Evaluation) counts the number of n-grams shared between the proposed and model summaries. The ROUGE toolkit [18] has been used to calculate ROUGE-1, ROUGE-2, ROUGE-L along with the precision and recall values. The results have been summarized below in a tabular form in tables 1, 2 and 3. We have also compared our approach to state of the art graph based approaches such as TextRank and LexRank. The results for the same have been summarized in table 4 and table 5.

| DUC 2002 | | | |
|---|---|---|---|
| Metric | Recall | Precision | F-Score |
| Rouge-1 | 0.447 | 0.239 | 0.311 |
| Rouge-2 | 0.163 | 0.087 | 0.114 |
| Rouge-L | 0.406 | 0.218 | 0.284 |

Table **1.** ROUGE metric scores for DUC 2002 using proposed method.

| 100 DOCUMENT DATASET | | | |
|---|---|---|---|
| Metric | Recall | Precision | F-Score |
| Rouge-1 | 0.443 | 0.626 | 0.519 |
| Rouge-2 | 0.324 | 0.495 | 0.392 |
| Rouge-L | 0.427 | 0.603 | 0.500 |

Table **2.** ROUGE metric scores for 100 Documents dataset using proposed method.

| 25 Documents from CNN/Daily Mail news | | | |
|---|---|---|---|
| Metric | Recall | Precision | F-Score |
| Rouge-1 | 0.491 | 0.138 | 0.216 |
| Rouge-2 | 0.132 | 0.031 | 0.051 |
| Rouge-L | 0.449 | 0.124 | 0.195 |

Table **3.** ROUGE metric scores for CNN/Daily Mail news using proposed method.

| DUC 2002 | | | |
|---|---|---|---|
| Method | Rouge-1 | Rouge-2 | Rouge-L |
| TextRank | 0.360 | 0.106 | 0.352 |
| LexRank | 0.373 | 0.214 | 0.368 |
| Our Method | 0.311 | 0.114 | 0.284 |

Table **4.** ROUGE metric scores comparison against traditional graph-based methods for DUC 2002

| 100 DOCUMENT DATASET | | | |
|---|---|---|---|
| Method | Rouge-1 | Rouge-2 | Rouge-L |
| TextRank | 0.567 | 0.439 | 0.546 |
| LexRank | 0.543 | 0.419 | 0.531 |
| Our Method | 0.519 | 0.392 | 0.501 |

Table **5.** ROUGE metric scores comparison against traditional graph-based methods for 100 document datasets

## 6. CONCLUSION

This paper explores the use of Hypergraphs and its ability to group each sentence to more than one group of topics, thereby increasing topical coverage in the summary and consequently improving its quality. For this study we have tested our approach on single document text summarization using keyword extraction and summary generation using the property of vertex domination. Based on the results we can observe that our method gives encouraging results which are comparable to the scores of TextRank and LexRank, especially in the case of the ESSG dataset [16] and proves to be a good alternative to other graph based methods. The low ROUGE scores for DUC 2002 and CNN/Daily_Mail dataset can be attributed to the abstractive nature of the reference summaries due to which there aren't many overlapping or common words between the reference and generated summary. However, our approach does not require any sentence scoring and sentence ranking computation thereby making it easier to implement and computationally effective. Further improvements in the ROUGE scores can be brought about by implementing co-reference resolution, with the help of which we can find all expressions that refer to the same entity in a text. Our future work involves implementing this approach to multi-document query oriented summarization and determine the quality of summaries generated against state of the art graph based approaches.